\documentclass{article}

\usepackage{microtype}
\usepackage{graphicx}
\usepackage{subfigure}
\usepackage{booktabs} 
\usepackage[authoryear]{natbib}

\usepackage{hyperref}

\usepackage[accepted]{icml2024}

\usepackage{amsmath}
\usepackage{amssymb}
\usepackage{mathtools}
\usepackage{amsthm}
\usepackage[capitalize,noabbrev]{cleveref}
\usepackage{makecell}

\theoremstyle{plain}

\theoremstyle{definition}

\theoremstyle{remark}

\usepackage[textsize=tiny]{todonotes}
\usepackage[capitalize,noabbrev]{cleveref}
\usepackage{framed}
\usepackage{multirow}
\usepackage{svg}
\usepackage{listings}
\usepackage{xcolor}
\usepackage{wrapfig}
\usepackage{fdsymbol}
\usepackage{makecell}
\usepackage{booktabs}
\usepackage{rotating}
\usepackage{xcolor}
\definecolor{mygray}{RGB}{192,192,192}

\usepackage[most]{tcolorbox}
\usepackage{float}
\usepackage{xspace}
\tcbset{
  aibox/.style={
    width=\textwidth,
    top=10pt,
    colback=white,
    colframe=black,
    colbacktitle=black,
    enhanced,
    center,
    attach boxed title to top left={yshift=-0.1in,xshift=0.15in},
    boxed title style={boxrule=0pt,colframe=white,},
  }
}
\newtcolorbox{AIbox}[2][]{aibox,title=#2,#1}

\newtcolorbox{exmpbox}[3][]{
  colback=green2!5,
  colframe=blue1,
  fonttitle=\bfseries,
  left=.02in,
  right=.02in,
  bottom=.02in,
  top=.02in,
  title={#2},
  #1
}

\newcommand\our{MathScale}
\newcommand\ourmodel{\emph{MathScale}}
\icmltitlerunning{\our{}: Scaling Instruction Tuning for Mathematical  Reasoning}

\begin{document}

\twocolumn[
\icmltitle{\our{}: Scaling Instruction Tuning for Mathematical  Reasoning}



\icmlsetsymbol{equal}{*}

\begin{icmlauthorlist}
\icmlauthor{Zhengyang Tang}{cuhksz,msra,sribd}
\icmlauthor{Xingxing Zhang}{msra}
\icmlauthor{Benyou Wang}{cuhksz,sribd}
\icmlauthor{Furu Wei}{msra}
\end{icmlauthorlist}

\icmlaffiliation{cuhksz}{The Chinese University of Hong Kong, Shenzhen, China}
\icmlaffiliation{msra}{Microsoft Research Asia, Beijing, China}
\icmlaffiliation{sribd}{Shenzhen Research Institute of Big Data, Shenzhen, China}



\icmlkeywords{Large Language Models, Mathematical Reasoning, Instruction Tuning, Synthetic Data, ICML}

\vskip 0.3in
]

\printAffiliationsAndNotice{}  

\begin{abstract}
Large language models (LLMs) have demonstrated remarkable capabilities in problem-solving. However, their proficiency in solving mathematical problems remains inadequate. 
We propose \ourmodel{}, a simple and scalable method to create high-quality mathematical reasoning data using frontier LLMs (e.g., {\tt GPT-3.5}).
Inspired by the cognitive mechanism in human mathematical learning, it first extracts topics and knowledge points from seed math questions and then build a concept graph, which is subsequently used to generate new math questions.
\ourmodel{} exhibits effective scalability along the size axis of the math dataset that we generate. As a result, we create a mathematical reasoning dataset (MathScaleQA) containing two million math question-answer pairs.
To evaluate mathematical reasoning abilities of LLMs comprehensively, we construct {\sc MwpBench}, a benchmark of Math Word Problems, which is a collection of ten datasets (including GSM8K and MATH) covering K-12, college, and competition level math problems. 
We apply MathScaleQA to fine-tune open-source LLMs (e.g., LLaMA-2 and Mistral), resulting in significantly improved capabilities in mathematical reasoning. 
Evaluated on {\sc MwpBench}, \our{}-7B achieves state-of-the-art performance across all datasets, surpassing its best peers of equivalent size by 42.9\% in micro average accuracy and 43.7\% in macro average accuracy, respectively.
\end{abstract}

\section{Introduction}
\label{sec:Introduction}
Large language models (LLMs) have demonstrated remarkable capabilities in problem-solving. However, their proficiency in solving mathematical problems remains inadequate, potentially due to the inherent necessity for multi-step complex reasoning in mathematical problem-solving. Instruction Tuning \cite{wei2021finetuned} is an effective approach to unlock certain capabilities in LLMs. Unfortunately, this approach is constrained by the limited size of the currently available datasets on mathematical reasoning. For example, the most popular math datasets, GSM8K \cite{gsm8k} and MATH \cite{MATH}, each only contains around 7.5K training examples.

An effective method to tackle this challenge is to augment existing high-quality math datasets using frontier LLMs such as {\tt GPT-3.5} and {\tt GPT-4}. For instance, WizardMath \cite{luo2023wizardmath} introduces an array of operations for {\tt GPT-3.5} to generate math questions with increased complexity. MetaMath \cite{yu2023metamath} bootstraps questions in GSM8K and MATH through answer augmentation, question rephrasing, self-verification and FOBAR questions. The newly generated examples by these methods exhibit substantial similarity to the original examples contained within the training set, which limits their power in generating large scale math datasets.

We therefore propose a conceptually simple and scalable method \ourmodel{}, which is less dependent on original training examples. Specifically, we first prompt {\tt GPT-3.5} to extract high level concepts (i.e., topics and knowledge points) from existing seed math questions. In this step, we convert concrete math questions to extractions and the dependency to original questions is largely removed. Given these extractions, we then build a concept graph, which is used to estimate the connections between different concepts. Finally, we can instruct {\tt GPT-3.5} to generate new math questions based on randomly sampled concepts from the graph. Intuitively, we can generate significantly more examples using different combination of concepts than using augmentation-based methods, since the resulting number of new examples is bounded by the number of augmentation operations. \ourmodel{} also bears resemblance to the cognitive mechanisms underlying the process of mathematical learning in humans \cite{humanlearn}. \citet{humanlearn} argues that the learning process of human involves two distinct steps called \emph{concept compression} and \emph{connection forging}. \emph{Concept compression} mirrors the process of high level concept extraction, while \emph{connection forging} is similar to our concept graph construction. 

Mathematical capability evaluation is another issue arising from the lack of high-quality mathematical datasets. Recently, most LLMs employ GSM8K \cite{gsm8k} and MATH \cite{MATH} for evaluation. However, GSM8K focuses on elementary-level problems, while MATH offers competition-level challenges. There is a clear gap between the two kinds of capabilities measured. 
Therefore, we introduce {\sc MwpBench}, a comprehensive and unified benchmark to measure mathematical reasoning capabilities. {\sc MwpBench} is composed of ten different math word problem datasets (including GSM8K and MATH) and it covers math word problems from elementary school to college level with different difficulty levels. Moreover, {\sc MwpBench} standardizes evaluations across all datasets with a unified protocol, promoting consistent and fair model comparisons.

\ourmodel{} exhibits effective scalability along the size axis of the math dataset that we generate. As a result, we create a mathematical reasoning dataset (MathScaleQA) containing two million math question-answer pairs. We apply MathScaleQA to fine-tune open-source LLMs (e.g., LLaMA-2 and Mistral), resulting in significantly improved capabilities in mathematical reasoning. Evaluated on {\sc MwpBench}, \our{}-7B achieves 35.0\% in micro average accuracy and 37.5\% in macro accuracy, outperforming its best peers of equivalent size by 42.9\% and 43.7\%, respectively.

\section{{\sc MwpBench} Evaluation Framework}
\label{sec:evluation}

\subsection{{\sc MwpBench}}

\textbf{Existing Datasets} Our first endeavor is to collate established datasets, including GSM8K~\cite{gsm8k}, MATH~\cite{MATH}, TAL-SCQ~\cite{TAL-SCQ}, Math23k~\cite{Math23k}, Ape210k~\cite{Ape210k}, GaokaoBench-Math~\cite{GaokaoBench}, and AGIEval~\cite{AGIEval} series (see Table \ref{tab:all_data_table}). Types of problems of these datasets are different. For example, most datasets contain math word problems, while TAL-SCQ comprises multi-choice questions. Intuitively, multi-choice questions are simpler because LLMs only need to figure out which choice leads to a higher probability. Therefore, we convert all multi-choice questions to math word problems (detailed in Appendix \ref{appendix:ToWordProblem}).
Secondly, some of the datasets (e.g., Math23k, Ape210k) are not in English and we translate them to English to expand existing math datasets (detailed in Appendix \ref{appendix:ToEnglishProblem}). Note that we translated part of their training sets and full test sets into English.

\begin{table*}[h!]
\centering
\footnotesize
\begin{tabular}{lllllcc}
\toprule
 \textbf{Dataset} & \textbf{Level} & {\bf Difficulty} & \textbf{Question Type} & \textbf{Language} & \textbf{\#Train} & \textbf{\#Test}  \\
\midrule
GSM8K & Elementary & Easy & Word  & En & 7473 & 1319 \\
MATH & Competition & ExHard & Word  & En & 7498 & 5000 \\
TAL-SCQ & K12 Math & Medium & MC$\rightarrow$Word & En & 2638 & 1496 \\
Math23k & Elementary & Easy & Word & Zh$\rightarrow$En & 1000 & 949 \\
Ape210k & Elementary & Easy & Word & Zh$\rightarrow$En & 967 & 4874 \\
GaokaoBench-Math & High School & Hard & MC$\rightarrow$Word & Zh$\rightarrow$En & 0 & 508 \\
AGIEval-Gaokao-Math & High School & Hard & MC$\rightarrow$Word & Zh$\rightarrow$En & 0 & 404 \\
AGIEval-SAT-Math & High School & Hard & MC$\rightarrow$Word & En & 0 & 102 \\
AGIEval-Math & Competition & ExHard & Word & En & 0 & 938 \\
CollegeMath & College & ExHard & Word & En & 1281 & 2818 \\
\midrule
Total & -- & -- & -- & -- & 20857 & 18408 \\
\bottomrule
\end{tabular}
\caption{Statistics in {\sc MwpBench}. In the ``Question Type'' column, ``Word'' stands for math word problem and ``MC'' stands for multiple-choice problem. In the ``Difficulty'' column, ``ExHard'' stands for extremely hard.}
\label{tab:all_data_table}
\end{table*}

\textbf{CollegeMath} Existing datasets does not cover college-level mathematics which requires diverse skills such as analytical thinking, logical reasoning, and quantitative analysis. We therefore propose CollegeMath to bridge this gap.

We curated a collection of nine college mathematics textbooks, each addressing a distinct topic (see Table \ref{tab:CollegeMathSpecifics} for more details). These textbooks encompass seven critical mathematical disciplines: algebra, pre-calculus, calculus, vector calculus, probability, linear algebra, and differential equations. These textbooks are originally in PDF format and we convert them to text format using the Mathpix API\footnote{\url{https://docs.mathpix.com/\#process-a-pdf}}, where equations are transformed to LaTeX format. Once converted a textbook to text format, we are ready to extract exercises and their solutions. For each book, we first manually segment the book into chapter and identify pages with exercises and their solutions. Then we extract questions in exercises and their associated short answers (see more details of our prompts in Appendix \ref{appendix:Extract}). In total, this dataset contains 1281 examples for training and 2818 examples for test.

\begin{table*}[h!]
\centering
\footnotesize
\begin{tabular}{ccccc}
\toprule
 \textbf{Topic} & \textbf{Book} & \textbf{License} & \textbf{\#Train} & \textbf{\#Test}  \\
\midrule
\makecell[l]{Algebra} & \makecell[l]{Beginning and Intermediate Algebra~\cite{book_algebra}} & \makecell[l]{CC BY 3.0} & 1171 & 1000 \\
\makecell[l]{Precalculus} & \makecell[l]{PRECALCULUS~\cite{book_precalculus}} & \makecell[l]{CC} & 80 & 500 \\
\makecell[l]{Calculus} & \makecell[l]{Calculus~\cite{book_calculus}} & \makecell[l]{CC BY-NC-SA} & 30 & 500 \\
\makecell[l]{VectorCalculus} & \makecell[l]{CORRAL's VECTOR CALCULUS~\cite{book_vector_calculus}} & \makecell[l]{GFDL} & 0 & 110 \\
\makecell[l]{Probability} & \makecell[l]{Introduction to Probability~\cite{book_probability_grinstead}} & \makecell[l]{GFDL} & 0 & 38 \\
\makecell[l]{Probability} & \makecell[l]{Probability and Statistics:\\The Science of Uncertainty~\cite{book_probability_evans}} & \makecell[l]{Custom\footnotemark} & 0 & 101 \\
\makecell[l]{LinearAlgebra} & \makecell[l]{Matrix Theory and LINEAR ALGEBRA~\cite{book_linear_algebra_peter}} & \makecell[l]{CC BY} & 0 & 123 \\
\makecell[l]{LinearAlgebra} & \makecell[l]{A First Course in LINEAR ALGEBRA~\cite{book_linear_algebra_kut}} & \makecell[l]{CC BY} & 0 & 137 \\
\makecell[l]{DifferentialEquations} & \makecell[l]{ELEMENTARY DIFFERENTIAL EQUATIONS~\cite{book_diff}} & \makecell[l]{CC BY-NC-SA 3.0} & 0 & 309 \\
\bottomrule
\end{tabular}
\caption{Details of permissively
licensed books we use to construct the CollegeMath dataset.}
\label{tab:CollegeMathSpecifics}
\vspace{-15pt}
\end{table*}

\footnotetext{Copyright (c) by Michael J. Evans and Jeffrey S. Rosenthal. It may be copied and distributed without restriction, provided it is not altered, appropriate attribution is given and no money is charged.}

\subsection{Unified Evaluation Protocol}
One of the challenges in benchmarking LLMs for mathematical reasoning is the inconsistency across evaluation metrics and protocols used in different work~\cite{touvron2023llama2,luo2023wizardmath,yue2023mammoth}.

{\sc MwpBench} aims to evaluate the mathematical reasoning abilities of instruction tuned LLMs using a unified evaluation protocol. We employ zero-shot setting for evaluation and use the accuracy metric. The reason behind that is we believe fine-tuned LLMs should be able to answer questions directly without demonstrations, while in few-shot setting the final results may change with different set of demonstrations. For prompt template, we choose the Alpaca template~\cite{alpaca} as default, which is the most widely used for instruction tuning \cite{alpaca,luo2023wizardmath,yu2023metamath}. However, we support customized template just in case that LLMs are trained with a different instruction template (e.g., OpenAI ChatGPT template). For decoding, we choose greedy decoding to eliminate randomness in comparisons, selecting the top-1 completion as the solution. To further standardize the evaluation, we carefully implemented the answer extraction and verification processes (with high precision fuzzy match). 

We plan to open-source our evaluation framework.

\section{\our{}: Scaling Instruction Tuning for Mathematical Reasoning}
\label{sec:Model}

\begin{figure*}[h]
\centering
\includegraphics[width=1.0\textwidth]{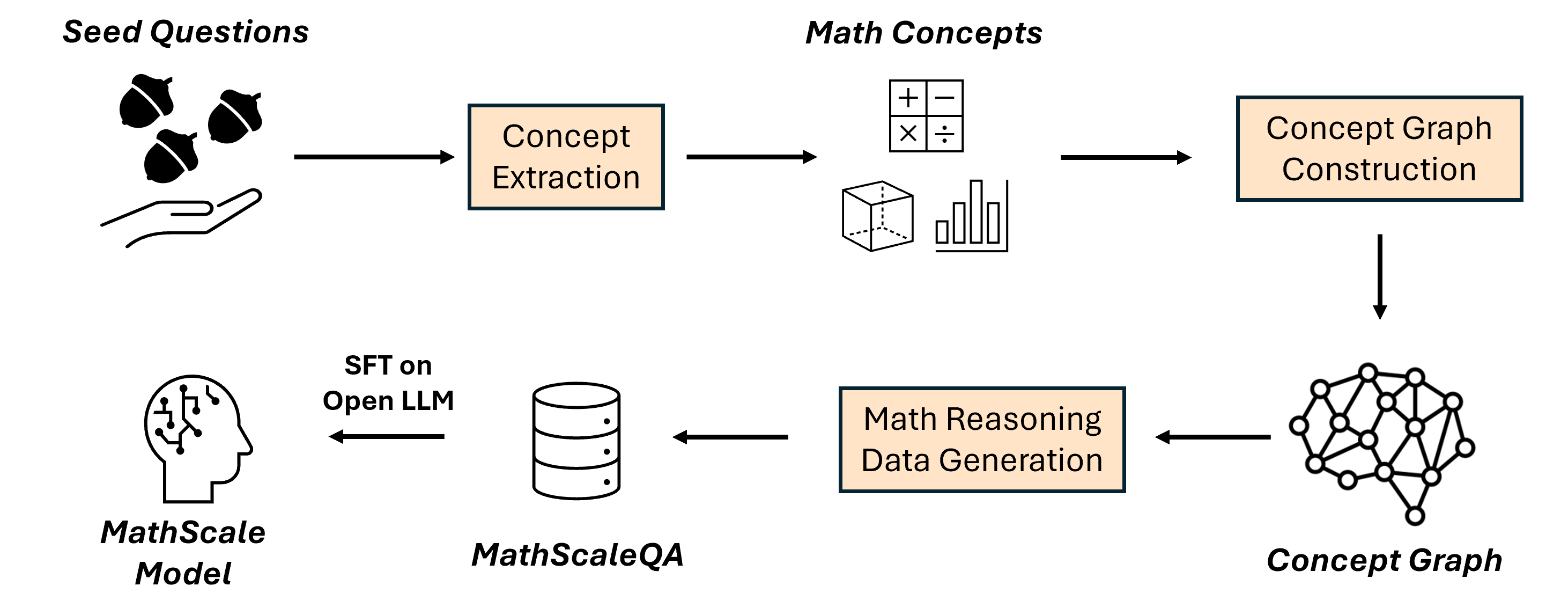}
\caption{Overview of {\our{}}. \our{} starts from seed math questions and there are three steps in this pipeline (i.e., \emph{concept extract}, \emph{concept graph construction} and \emph{mathematical reasoning data generation}). After these three steps, we obtain the MathScaleQA dataset, which is subsequently used to train open LLMs. Finally, we obtain \our{} models.} 
\label{fig:method_overview}
\end{figure*}

We present details of \our{} in this section. \our{} aims to generate large scale Mathematical Reasoning dataset by prompting ChatGPT and it contains four steps.

\subsection{Concept Extraction}
\label{sec:concept_extraction}
As shown in Figure \ref{fig:method_overview}, \our{} takes seed math questions as input and we use the training set of {\sc MwpBench} (around 20K math questions). In the first step, we extract high level concepts (i.e., topics and knowledge points) from these seed questions with prompt engineering of {\tt GPT-3.5}. We aim to extract meta information needed to solve a particular math question. We believe ``topics'' and ``knowledge points'' are important meta information for questions. A ``topic'' refers to the mathematical subject name or the topic name of math book chapter such as ``Money and finance'' and ``Arithmetic operations''. While ``knowledge points'' refers to more fine grained math concepts (e.g., theorems, skills) in problem solving. Typical examples are ``Definition and properties of dot product'' or ``Converting fractions to whole numbers''. We instruct {\tt GPT-3.5} to act as a Math teacher and extract 1 or 2 topics and 1 to 5 knowledge points from a given seed question (see the prompt template in Table \ref{tbl:concept}).

\begin{table}[h!]
\centering
\small
\begin{tabular}{|l|}
\hline
\makecell[l]{\\Act as a Math Teacher and analyze the provided question. \\
Start by identifying 1 or 2 general topics that a student is\\ being assessed on. Next, highlight 1 to 5 specific knowledge\\ points that the question evaluates. \\
\\
Provided question: {\tt \{seed\_question\}} \\
\\
Analysis:\\
\\
} \\
\hline
\end{tabular}
\caption{Prompt template for Concept Extraction.}
\label{tbl:concept}
\vspace{-15pt}
\end{table}

To ensure the diversity of the extracted topics and knowledge points, we use the training set of {\sc MwpBench}, which includes questions from different sources. We also remove topics and knowlege points that appear only one time to reduce noise. In total, we extracted around 2K topics and 8K knowledge points. The above process mirrors the \emph{concept compression} described in \cite{humanlearn}.

\subsection{Concept Graph Construction}
\label{sec:concept_graph}
\begin{figure*}[h]
\centering
\includegraphics[width=0.9\textwidth]{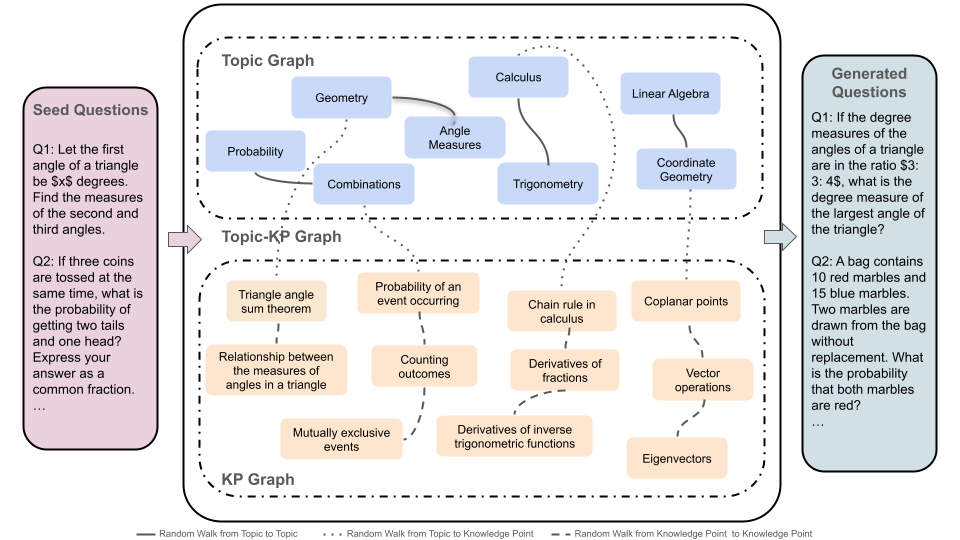}
\caption{Running Examples of the concept graph construction process in the \our{} pipeline.}
\label{fig:runningexample}
\vspace{-15pt}
\end{figure*}

\textbf{Concept Graph} Given the topics and knowledge points extracted from the previous step, we move on to construct a concept graph $C$, whose nodes are the extracted topics $\mathbb{T}=\{\mathbf{t}_1, \mathbf{t}_2, \dots, \mathbf{t}_{|\mathbb{T}|} \}$ and knowledge points (KPs) $\mathbb{K}=\{\mathbf{k}_1, \mathbf{k}_2, \dots, \mathbf{k}_{|\mathbb{K}|} \}$. As shown in Figure \ref{fig:runningexample}, we have three types of edges in this graph (i.e., topic to topic edge, topic to KP edge and KP to KP edge), which results to three sub-graphs (topic graph, topic-KP graph, KP graph). When a topic (or KP) $\mathbf{u}$ is co-occurred with another topic (or KP) $\mathbf{v}$, we build an edge between them and the edge weight is related to their co-occurrence statistics. Define co-occurrence as $\mathbf{u}$ and $\mathbf{v}$ have been extracted from the seed question. 

Formally, let $E=\{ (\mathbf{u}, \mathbf{v}) | f_{\text{co}}(\mathbf{u}, \mathbf{v}) > 0 \}$ denote edges in $C$ and $f_{\text{co}}(\mathbf{u}, \mathbf{v})$ is the edge weight between $\mathbf{u}$ and $\mathbf{v}$. Intuitively, two KPs (or topics) are more likely to be reasonable composition when they have been frequently used to solve the same seed questions. Let \( w_{\mathbf{uv}} \) denote the raw co-occurrence count between node $\mathbf{u}$ and node $\mathbf{v}$. The adjusted weight $f_{\text{co}}(\mathbf{u}, \mathbf{v})$ is defined as follows:
\begin{equation}
\label{eq:edge_weight}
    f_{\text{co}}(\mathbf{u}, \mathbf{v}) = \log( w_{\mathbf{uv}} + \varepsilon)
\end{equation}
where \( \varepsilon \) is a small constant introduced to maintain non-zero counts and prevent computational issues. 

\textbf{Concept Composition} Given the graph $C$, we are ready to sample topics and KPs from it and the sampled topics and KPs are subsequently used to generate new math questions. We use a graph random walk algorithm to create concept compositions. 

We start from a uniformly random sampling from the $|\mathbb{T}|$ topics we have extracted. Note that in implementation, we simply enumerate all extracted topics for multiple epochs. 

In the second step, we do a random walk for one to two steps in the topic sub-graph to search for related topics. The probability distribution for the graph random walk is not uniform and defined as follows:
\begin{equation}
    \label{eq:probs}
    p_{\mathbf{uv}} = \frac{\exp(f_{\text{co}}(\mathbf{u}, \mathbf{v}))}{\sum_{\mathbf{v}' \in \mathcal{N}(\mathbf{u})} \exp(f_{\text{co}}(\mathbf{u}, \mathbf{v}'))}
\end{equation}
where $\mathcal{N}(\mathbf{u})$ denotes the set of nodes adjacent to $\mathbf{u}$ in the {\bf topic} sub-graph.

In the third step, we continue to randomly walk in the hybrid topic-KP graph for a single step with the probability distribution calculated as in Equation (\ref{eq:probs}) on the {\bf topic-KP} graph. So that we now have one sampled KP. 

In the last step, we continue to expand to more KPs by randomly walking on the KP graph for zero to four steps again with the probability distribution computed as in Equation (\ref{eq:probs}) on {\bf KP} graph. We finally obtained a set of sampled topics $\hat{\mathbb{T}}$ and KPs $\hat{\mathbb{K}}$.

The whole process above is an analogy of the \emph{connection forging} described in \cite{humanlearn}.

\subsection{Mathematical Reasoning Data Generation}

\begin{table}[h]
\centering
\small
\begin{tabular}{|l|}
\hline
\makecell[l]{\\Act as a Math Teacher and create a new question and its solution\\ based on the provided topics and knowledge points. Ensure that \\the created questions: \\
\\
1. Adhere to the provided topics.\\
2. Necessitate the combined use of the associated knowledge\\
points.\\
\\
{\tt \{few\_shot\_examples\}}\\
\\
Topics:\\
{\tt \{topics\}}\\
\\
Knowledge Points:\\
{\tt \{knowledge\_points\}}\\
\\
Structure your response as:\\
FORMAT INSTRUCTIONS OF THE NEW QA-PAIR ...\\
\\
} \\
\hline
\end{tabular}
\caption{Prompt template for Mathematical Reasoning Data Generation.}
\label{tbl:generte}
\vspace{-15pt}
\end{table}

With the novel compositions of topics $\hat{\mathbb{T}}$ and KPs $\hat{\mathbb{K}}$ at hand, we query {\tt GPT-3.5} to generate corresponding question-answer pairs. Inspired by how math teachers design questions from existing exercises, we opt to include few-shot examples to guide {\tt GPT-3.5} in question formulation. These examples are chosen from the seed questions, based on the Jaccard distance of their knowledge points set. We ask {\tt GPT-3.5} to adhere to $\hat{\mathbb{T}}$ and encourage combine use of KPs $\hat{\mathbb{K}}$. We present the template for prompts in Table \ref{tbl:generte}.

Furthermore, we apply a decontamination process, where all math questions in the test set of {\sc MwpBench} are removed.

\subsection{Validation}
\label{sec:valid_method}

We observe that sometimes in the newly generated QA pairs, the solution is incorrect. We therefore also tried to add an additional validation process as follows. We first instruction {\tt GPT-4} to generate a reference solution for the question and then ask {\tt GPT-4} again to validate the {\tt GPT-4} solution against the solution generated in the previous step. We assume {\tt GPT-4} is more accurate than {\tt GPT-3.5}. If {\tt GPT-4} believe the orignal solution is incorrect, we replace it with the new {\tt GPT-4} solution.  Small scale experiments (Table \ref{tab:validation}) show the step does not improve the results. Perhaps because essentially we are trying to distill {\tt GPT-3.5} using open source LLMs. Although some solutions are incorrect, they are still help open source LLMs to learn the model distributions of {\tt GPT-3.5}.
Therefore, in our final pipeline, we remove this validation step.

\section{Experiments}
\label{sec:Experiment}

\subsection{Implementation}
\label{sec:impl}
\paragraph{Data Generation} In concept extraction (Section \ref{sec:concept_extraction}), we use the {\sc MwpBench} training set, comprising around 20K questions, as the seed questions for our \our{} pipeline and we employ {\tt GPT-3.5-Turbo-0613} for the extraction. In total, we obtain 2,018 topics and 8,892 knowledge points.
We then construct graphs to establish relationships among these concepts (Section \ref{sec:concept_graph}). The edge weight in the graph is smoothed using Equation (\ref{eq:edge_weight}) and we set $\varepsilon=1e-5$.
In the concept composition process, treating the iteration through all topic nodes as one epoch, we repeat this process for approximately 1K epochs, resulting 2 million unique concept compositions. Then we instruct {\tt GPT-3.5-Turbo-0613} to create 2 million question-answer pairs with these compositions. We also decontaminate the generated datasets by excluding all math questions in the test set of {\sc MwpBench}. To leverage the precious high quality math reasoning data, we additionally combine the generated data with the training set of {\sc MwpBench}. We call the resulting dataset {\bf MathScaleQA}. The validation step (Section \ref{sec:valid_method}) is excluded from the final pipeline, because we find that the validation step does not improve results (see details in Section \ref{sec:validation}). Example outputs for each step of the pipeline are provided in Appendix \ref{appendix:concrete_examples}.

\paragraph{Model Training}  
The questions in MathScaleQA are formatted using the Alpaca prompt \cite{alpaca} as follows. 

{\tt Below is an instruction that describes a task. Write a response that appropriately completes the request.\\
\\
\#\#\# Instruction: \\
\{question\} \\
\\
\#\#\# Response:}

Our training pipeline is adapted from the open-instruct \cite{tulu} toolkit. We utilize the LLaMA-2 7B and 13B models~\cite{touvron2023llama2} as well as the Mistral 7B model \cite{jiang2023mistral} as our backbone models.
We use a batch size of 128 and train on the MathScaleQA dataset for 3 epochs using a learning rate of 2e-5. We call the resulting models \ourmodel{}-7B, \ourmodel{}-13B and \ourmodel{}-\emph{Mistral-7B}. We leave exploration of the LLaMA-2 70B model in future work.

\subsection{Models in Comparison}
For a comprehensive evaluation, we select a diverse set of previous LLMs specialized in mathematical reasoning for comparison.

\paragraph{Close-Source Models} We include the most capable GPT models developed by OpenAI, which are the light-weighted {\tt GPT-3.5-Turbo-0613} and the powerful {\tt GPT-4-0314}. These models are known to be good at mathematical reasoning and serves as the upper bounds.

\textbf{Open-Source Models:} We also compare our model against open-source math models. Specially, we compare with WizardMath~\cite{luo2023wizardmath}, GAIR-Abel~\cite{abel}, MetaMath~\cite{yu2023metamath}, and MAmmoTH~\cite{yue2023mammoth}. WizardMath~\cite{luo2023wizardmath} is based on evol-instruct \cite{xu2023wizardlm}  and reinforcement learning.
MetaMath~\cite{yu2023metamath} is trained on a dataset by augmenting GSM8K \cite{gsm8k} and MATH \cite{MATH} using answer or question side paraphrasing. The dataset used to train MAmmoTH~\cite{yue2023mammoth} comprises a collection of 13 existing math datasets with GPT-4 CoT \cite{wei2022chain} and/or PoT \cite{pmlr-v202-gao23f,chen2022program} annotations. We evaluate all models using CoT natural language style math solutions. We noticed that some of the models (e.g., {\tt GPT-4} and MAmmoTH) can produce code solution of math problems in addition to natural language solutions. For fair comparison, we refrain from comparing using code-interpreter style solutions, because all models above can produce code-interpreter style solutions if the solutions in their training data are replace by {\tt GPT} annotated code solutions. Also note that WizardMath v1.1 is a Mistral based math model and we do not know how its training data are constructed (the authors did not release any detail of the training data of WizardMath v1.1).
We evaluate all models on {\sc MwpBench}, which contains 10 datasets on mathematical reasoning. We report accuracies of the 10 datasets as well as their micro-average and macro-average. We prompt all models using the Alpaca template (see Section \ref{sec:impl}). \cite{luo2023wizardmath} recommended an improved prompt for during inference (i.e., adding {\tt Let's think step by step} after the standard Alpaca template). However, we observe mixed results on {\sc MwpBench} for some models in comparison. For example, we observe improved results on GSM8K, but decreased results on MATH. We therefore do not use this optimization for all models in comparison.

\subsection{Main Results}

\begin{table*}[h!]
\centering
\hspace{-2.8mm}
\footnotesize
\setlength{\tabcolsep}{3pt}
\renewcommand{\arraystretch}{1.15}
\begin{tabular}{lcccccccccccc}
\toprule
\textbf{Models} & \textbf{GSM8K} & \textbf{MATH} & \makecell[l]{\textbf{College}\\\textbf{Math}} & \textbf{TAL} & \makecell[l]{\textbf{Math23k}} & \makecell[l]{\textbf{Ape210k}} & \makecell[l]{\textbf{Gaokao}\\\textbf{Bench}\\\textbf{Math}} & \makecell[l]{\textbf{AGIE}\\\textbf{Gaokao}\\\textbf{Math}} & \makecell[l]{\textbf{AGIE}\\\textbf{SAT}\\\textbf{Math}} & \makecell[l]{\textbf{AGIE}\\\textbf{MATH}} & \makecell[l]{\textbf{Micro}\\\textbf{Average}} & \makecell[l]{\textbf{Macro}\\\textbf{Average}}\\
\midrule
\multicolumn{13}{c}{\textit{Closed-source Models}} \\
\makecell[l]{GPT-4} & \textbf{92.9} & \textbf{51.8} & \textbf{24.4} & \textbf{51.8} & \textbf{76.5} & \textbf{61.5} & \textbf{35.4} & \textbf{28.2} & \textbf{68.6} & \textbf{50.7} & \textbf{52.0} & \textbf{54.2} \\
\makecell[l]{GPT-3.5-Turbo} & 74.1 & 37.8 & 21.6 & 42.9 & 62.5 & 44.0 & 23.2 & 15.3 & 55.8 & 37.4 & 39.8 & 41.5\\
\midrule
\multicolumn{13}{c}{\textit{Models based on LLaMA-2 13B}} \\
\makecell[l]{LLaMA-2 13B} & 7.1 & 3.5 & 1.2 & 6.3 & 9.5 & 7.9 & 0.7 & 0.4 & 6.8 & 3.7 & 4.5 & 4.7 \\
\makecell[l]{WizardMath} & 62.0 & 14.3 & 7.8 & 18.7 & 38.3 & 25.2 & 8.2 & 3.4 & 29.4 & 15.8 & 20.2 & 22.3 \\
\makecell[l]{MAmmoTH} & 56.5 & 12.6 & 6.5 & 17.3 & 39.5 & 28.1 & 5.9 & 4.9 & 20.5 & 12.5 & 18.9 & 20.4 \\
\makecell[l]{GAIR-Abel} & 66.4 & 16.6 & 7.9 & 21.1 & 42.2 & 27.8 & 7.0 & 4.9 & 30.3 & 18.2 & 22.3 & 24.3 \\
\makecell[l]{MetaMath} & 70.8 & 22.8 & 10.1 & 25.4 & 48.6 & 31.6 & 9.6 & 5.6 & 38.2 & 22.9 & 26.8 & 28.6\\
\makecell[l]{\our{} 13B} & \textbf{71.3} & \textbf{33.8} & \textbf{20.4} & \textbf{38.1} & \textbf{61.1} & \textbf{43.7} & \textbf{20.0} & \textbf{12.3} & \textbf{55.8} & \textbf{34.7} & \textbf{37.1} & \textbf{39.1}\\
\midrule
\multicolumn{13}{c}{\textit{Models based on LLaMA-2 7B}} \\
\makecell[l]{LLaMA-2 7B} & 4.5 & 4.2 & 2.3 & 7.6 & 6.8 & 7.3 & 2.1 & 2.9 & 2.9 & 5.0 & 4.7 & 4.6 \\
\makecell[l]{WizardMath} & 52.8 & 10.3 & 6.8 & 14.0 & 32.5 & 19.2 & 5.9 & 6.1 & 22.5 & 11.7 & 15.8 & 17.1 \\
\makecell[l]{MAmmoTH} & 50.0 & 9.5 & 6.2 & 13.3 & 34.6 & 21.4 & 3.9 & 2.7 & 19.6 & 10.9 & 15.6 & 17.2 \\
\makecell[l]{GAIR-Abel} & 57.6 & 12.7 & 6.6 & 18.3 & 35.4 & 24.5 & 4.3 & 4.4 & 23.5 & 14.6 & 18.5 & 20.2 \\
\makecell[l]{MetaMath} & 66.2 & 20.6 & 9.4 & 22.5 & 44.0 & 29.9 & 5.9 & 5.1 & 36.2 & 20.8 & 24.5 & 26.1\\
\makecell[l]{\our{} 7B} & \textbf{66.3} & \textbf{31.1} & \textbf{20.9} & \textbf{35.2} & \textbf{59.0} & \textbf{41.8} & \textbf{19.6} & \textbf{12.6} & \textbf{57.8} & \textbf{31.1} & \textbf{35.0} & \textbf{37.5} \\
\midrule
\multicolumn{13}{c}{\textit{Models based on Mistral 7B}} \\
\makecell[l]{Mistral 7B} & 15.5 & 10.1 & 7.5 & 17.9 & 18.5 & 15.5 & 6.2 & 5.9 & 22.5 & 10.4 & 11.9 & 13.0 \\
\makecell[l]{WizardMath v1.1} & \textbf{78.1} & 32.8 & 16.0 & 34.4 & 58.3 & 41.4 & 16.1 & 9.6 & 55.8 & \textbf{33.0} & 35.4 & 37.6 \\
\makecell[l]{MetaMath Mistral} & 77.4 & 28.4 & 15.7 & 31.4 & 55.1 & 38.1 & 15.3 & 10.1 & 50.9 & 28.4 & 32.7 & 35.1 \\
\makecell[l]{\our{} Mistral} & 74.8 & \textbf{35.2} & \textbf{21.8} & \textbf{39.9} & \textbf{64.4} & \textbf{46.0} & \textbf{21.4} & \textbf{14.3} & \textbf{57.8} & 32.9 & \textbf{38.7} & \textbf{40.8} \\
\bottomrule
\end{tabular}
\caption{Performance metrics on {\sc MwpBench}. All evaluations were conducted utilizing the driver provided by {\sc MwpBench}, ensuring a consistent and fair comparison. Within each section, the highest results are highlighted in bold font. ``AGIE'' stands for AGIEval.}
\label{tab:main}
\vspace{-10pt}
\end{table*}

As shown in Table \ref{tab:main}, \our{} obtains best micro average and macro average scores on {\sc MwpBench} compared to other models based on LLaMA-2 7B, \mbox{LLaMA-2} 13B or \mbox{Mistral} 7B. Specifically, On average, \ourmodel{}-7B achieves a 35.0\% (micro) and 37.5\% (macro) accuracy across {\sc MwpBench}, surpassing its best counterparts of equivalent size by 42.9\% and 43.7\%, respectively. The trends are similar for \ourmodel{}-13B and \ourmodel{}-Mistral. This also confirms the effectiveness of our MathScaleQA dataset regardless of the backbone model.
Note that in GaokaoBench-Math, AGIEval-Gaokao-MATH, and AGIEval-SAT-MATH, there is no training set. Even on these out-of-domain test sets, \ourmodel{}-7B wildly outperforms other open-source models in comparison. When compared to frontier LLMs, \our{}-Mistral demonstrates performance parity in both micro and macro averages relative to {\tt GPT-3.5-Turbo} (see the first block in Table \ref{tab:main}).
We have also included subset performances on the MATH and CollegeMath datasets in Appendix \ref{sec:topics} to analyze model capabilities across different topics and disciplines.

\section{Analysis and Discussions}

\subsection{Scaling Property of \our{}}
\label{sec:scaling} 
\begin{figure*}[h]
\centering
\includegraphics[width=0.9\textwidth]{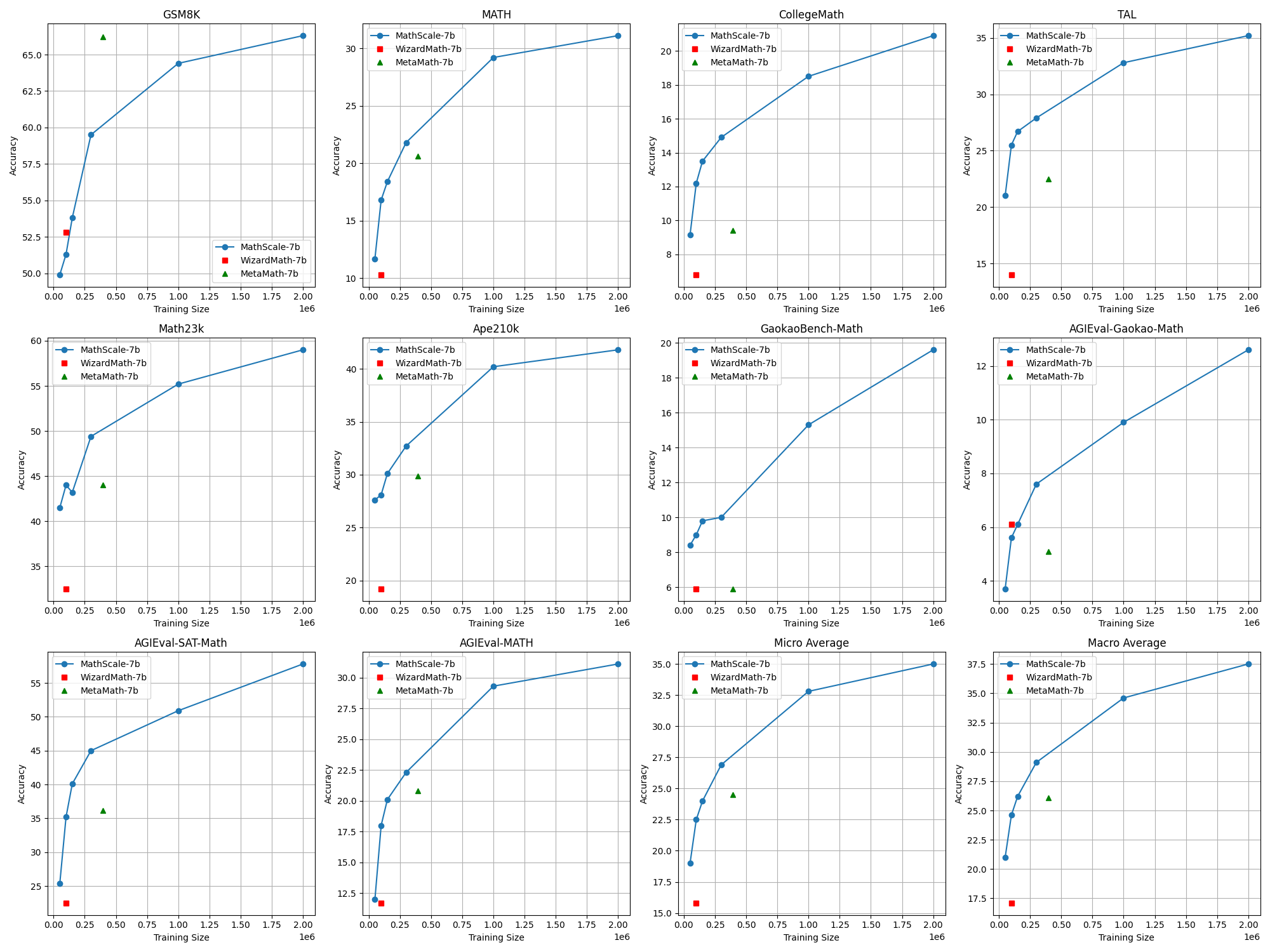}
\caption{Performance on {\sc MwpBench} using different sizes of training dataset in MathScaleQA.}
\label{fig:fix_kp.scale_train}
\vspace{-3pt}
\end{figure*}

As described in Section \ref{sec:Model},
given a fixed set of math concepts, iterating over concept graphs allows us to generate different compositions of mathematical concepts, thereby synthesizing large amount of new math data. We use LLaMA-2 7B as our base model to study the scaling property of \our{}.  When scaling the size of the MathScaleQA dataset, we observe a nearly logarithmic growth in the performance of the \our{}-7b model across all datasets within {\sc MwpBench}, as depicted in Figure \ref{fig:fix_kp.scale_train}. We draw the scaling curve up to two million examples (size of the full  MathScaleQA). We also compare \our{} against WizardMath and MetaMath at their respective training sizes. \our{} outperforms both models across all datasets (except for GSM8K) when using an equivalent amount of training data.
Given the scaling curves in Figure \ref{fig:fix_kp.scale_train}, we anticipate that the performance of \our{} may continue to improve with even more synthetic training examples. Due to resource constraints, we leave the training set scaling beyond two million examples to future work.

\subsection{Ablation on Concept Extraction}
In the concept extraction process (Section \ref{sec:concept_extraction}), we use all the 20K seed questions. We attempt to answer the following two questions. \emph{1) Does the number of seed questions matter?} \emph{2) Does the number of extracted concepts matter?} We control the size of resulting training examples to 25K for fast experimentation. In all experiments, we use the LLaMA-2 7B model as our backbone model.

\paragraph{Number of Seed Questions} To assess the influence of seed questions, we firstly randomly remove 50\% of the seed questions from the {\sc MwpBench} training set (i.e., we use only 10K seed questions). The results are shown in Table \ref{tab:fix_traininig_size}. We observe the macro average on {\sc MwpBench} drops by 2.9\%.  
Further, when we limite the data source of seed questions exclusively to the training sets of GSM8K and MATH, there is a performance decrease of 3.5\%. These results above indicate that incorporating of a larger and more diverse set of seed questions is beneficial.

\paragraph{Number of Math Concepts} Additionally, we examine the impact of extracted math concepts. As shown in Table \ref{tab:fix_traininig_size}, by removing half of the topics or knowledge points, we observe a notable decrease in the macro average on the {\sc MwpBench}. Particularly, removing knowledge points lead to a greater decrease in performance  (i.e., -8.6\% with 50\% knowledge points v.s. -2.3\% with 50\% of topics). This highlights the essential role that knowledge points play in enhancing the effectiveness of \our{}.

\begin{table}[h]
\centering
\footnotesize
\begin{tabular}{lcc}
\toprule
 \textbf{Methods} & \textbf{\makecell[c]{Macro\\Average}} & \textbf{\makecell[c]{Relative\\Change}}  \\
\midrule
\makecell[l]{\our{}} & 14.5 & - \\
\midrule
\makecell[l]{Remove 50\% Seed Questions} & 14.0 & -2.9\% \\
\makecell[l]{Restrict Data Source\\to GSM8K and MATH only} & 13.9 & -3.5\% \\
\midrule
\makecell[l]{Remove 50\% Topics} & 14.1 & -2.3\% \\
\makecell[l]{Remove 50\% Knowledge Points} & 13.2 & -8.6\% \\
\bottomrule
\end{tabular}
\caption{Ablation studies of concept extraction with a control training size of 25K on {\sc MwpBench}.}
\label{tab:fix_traininig_size}
\vspace{-10pt}
\end{table}

\subsection{On Validating Generated Data}
\label{sec:validation}

The generated QA pairs in MathScaleQA might be incorrect. Therefore, we introduce a separate validation step in Section \ref{sec:valid_method}. 
In this section, we design controlled experiment on 5K generated data from MathScaleQA and again using \mbox{LLaMA-2} 7B as our base model. 

\paragraph{GPT-4 v.s. GPT-3.5 Accuracy} We manually annotate 100 randomly chosen generated data points and generate answers with {\tt GPT-3.5-Turbo} and {\tt GPT-4}. {\tt GPT-4} demonstrate an impressive accuracy of 87\%, significantly outperforming the accuracy of 69\% by {\tt GPT-3.5-Turbo}. Therefore, we used {\tt GPT-4} to generate reference solutions and validate our synthetic solutions, replacing any incorrect solutions with the {\tt GPT-4} reference solutions.

\paragraph{Results} Within the 5K examples, 26\% of the solutions are identified as incorrect by {\tt GPT-4} and are replaced. We have another two settings with either all {\tt GPT-3.5} solutions and {\tt GPT-4} solutions. The results are shown in Table \ref{tab:validation} and we observe that using original {\tt 3.5-Turbo} solutions lead to a similar results as using the validation step.

This observation is counter-intuitive. Maybe because training on synthetic data generated from {\tt GPT-3.5} is essential distillation. Even if some solutions are incorrect, they may still help to the open-source LLMs (e.g., LLaMA-2 or Mistral) to mimic the distirubtions of {\tt GPT-3.5}. We also notice that in neural machine translation distillation, the step of validating incorrect translations is also ignored \cite{kim-rush-2016-sequence}.
Therefore, we opt to omit the validation and correction step from the final \our{} pipeline.

\begin{table*}[h!]
\centering
\footnotesize
\begin{tabular}{lcc}
\toprule
 \textbf{Methods} & \textbf{Micro Average} & \textbf{Macro Average}  \\
\midrule
\makecell[l]{100\% GPT-3.5 Solutions} & {\bf 10.6} & {\bf 11.5} \\
\makecell[l]{74\% GPT-3.5 Solutions and 26\% GPT-4 Corrected Solutions} & 10.2 & 11.1 \\
\makecell[l]{100\% GPT-4 Solutions} & 9.8 & 10.9 \\
\bottomrule
\end{tabular}
\caption{Ablation studies of validation step with a control training size of 5K on {\sc MwpBench}.}
\label{tab:validation}
\vspace{-10pt}
\end{table*}

\subsection{Performance on a Fresh Math Dataset}
\label{sec:leakage}

While MathScaleQA generated by {\tt GPT-3.5} is rigorously decontaminated to prevent overlap with the {\sc MwpBench} test set, there may still be small chance that some of the test sets have been leaked to {\tt GPT-3.5-Turbo} or contained in the training data of LLaMA-2. Because {\tt GPT-3.5-Turbo} uses human annotated queries submitted by users through their APIs\footnote{https://openai.com/research/instruction-following}. These queries may include test sets such GSM8K. The training set of LLaMA-2 is not released and we are not sure if some examples in test sets of {\sc MwpBench} are included or not.

To address this issue, we manually curate a new dataset comprising the latest 30 math questions from latest Gaokao Math exam, held in June for China National Higher Education Entrance Examination. We term this dataset, \emph{Fresh-GaokaoMath-2023}, which we believe Fresh-GaokaoMath-2023 is not likely to be included in the training data of LLaMA-2 or {\tt GPT-3.5-Turbo}. Because LLaMA-2 and {\tt GPT-3.5-Turbo} are released before Fresh-GaokaoMath-2023 is created.

We compare our LLaMA-2 7B based model \our{}-7B against two other LLaMA-2 7B based models (i.e., WizardMath-7B and  MetaMath-7B) as well as {\tt GPT-3.5-Turbo} and {\tt GPT-4}. Results are in Table \ref{tab:freshgaokao}. \our{} consistently surpasses  WizardMath and MetaMath, which aligns with the main results shown in Table \ref{tab:main}. It demonstrates the robustness and adaptability of \our{} in handling fresh math questions.

\begin{table}[h!]
\centering
\footnotesize
\begin{tabular}{lcc}
\toprule
 \textbf{Model} & \textbf{Fresh-GaokaoMath-2023} \\
\midrule
\makecell[l]{GPT-4} & 43.3 \\
\makecell[l]{GPT-3.5-Turbo} & 40.0 \\
\hline
\makecell[l]{WizardMath-7B} & 13.3 \\
\makecell[l]{MetaMath-7B} & 16.6 \\
\makecell[l]{\our{}-7B} & {\bf 30.0} \\
\bottomrule
\end{tabular}
\caption{Performance metrics on Fresh-GaokaoMath-2023.}
\label{tab:freshgaokao}
\vspace{-10pt}
\end{table}

\section{Related Work}
\textbf{ChatGPT-based Instruction Tuning} A pivotal aspect driving advancements in math instruction tuning is the use of ChatGPT for data synthesis. For instance, WizardMath~\cite{luo2023wizardmath} introduced reinforced evol-instruct which integrates five operations: adding constraints, deepening, concretizing, increasing reasoning steps, and complicating input, thereby facilitating comprehensive evolution. Similarly, MetaMath~\cite{yu2023metamath} employs a bootstrapping strategy for questions, incorporating answer augmentation, rephrasing, self-verification, and FOBAR. While these methods are effective, the breath space is inherently confined to manually designed operations. Our approach seeks to enable ChatGPT to emulate cognitive processes in human mathematical learning, thus overcoming the limitations faced by previous methodologies.

\textbf{Tool-Integration Instruction Tuning} Recent studies have also explored integrating tools into ChatGPT-based instruction tuning for mathematics. ToRA~\cite{tora} combines natural language reasoning with program-based tool usage to synthesize trajectory data. Each trajectory iteratively concatenates reasoning, programming, and program outputs until the final answer is reached. Our current focus is solely on natural language reasoning. While tool integration within the \our{} pipeline is an intriguing prospect, we reserve its exploration for future research.

\section{Conclusions}
We propose \ourmodel{}, a simple and scalable method to create high-quality mathematical reasoning data using frontier LLMs. We also construct {\sc MwpBench}, a comprehensive benchmark of Math Word Problems covering K-12, college, and competition level math problems. Evaluated on {\sc MwpBench}, \our{}-7B achieves state-of-the-art performance across all datasets, surpassing its best peers of equivalent size by 42.9\% in micro average accuracy and 43.7\% in macro average accuracy, respectively.

\section*{Broader Impact}

This paper seeks to advance mathematical reasoning by introducing a scalable method for generating high-quality synthetic data with large language models, along with new 
 evaluation benchmarks to foster consistent and fair model comparisons in academia. While our efforts center on assessing mathematical capabilities, it's crucial to note that the models may exhibit biases not examined in our study. Addressing these biases and ensuring the models' alignment with societal values is essential, highlighting the need for comprehensive evaluations that encompass both technical performance and ethical considerations.

\bibliography{example_paper}
\bibliographystyle{icml2024}

\newpage
\appendix
\onecolumn

\section{Appendix}\label{sec:appendix}

\subsection{{\sc MwpBench}: Transform Non-Word Problems into Word Problems}\label{appendix:ToWordProblem}
For datasets like TAL-SCQ~\cite{TAL-SCQ}, GaokaoBench-Math~\cite{GaokaoBench}, and AGIEval~\cite{AGIEval}, the problems are presented in a multiple-choice format. To eliminate the influence of the problem type and concentrate on the intrinsic ability of LLMs to address mathematical problems, we converted these non-word problems into word problems.

\subsubsection{Filtering Questions}
Initially, we identified and filtered out questions that rely heavily on the multiple-choice format. This filtering was done using specific keywords and phrases that are indicative of multiple-choice questions. 

\begin{lstlisting}[caption=Filtering questions]
def is_bad_question(question):
    question = question.lower()

    keywords = [
        "?",
        "which of the following",
        "which one",
        "which is",
        "the following",
        "which statement"
    ]
    
    for keyword in keywords:
        if keyword in question:
            print(f"Filtered question: {question}")
            return True
    return False
\end{lstlisting}

\subsubsection{Creating Question-Answer Pairs}
After filtering out the aforementioned questions, the remaining questions were paired with their corresponding correct answer choices. This transformation resulted in a format where each problem is presented as a word problem followed by its solution.

\subsection{{\sc MwpBench}: Translation of Non-English Problems to English}\label{appendix:ToEnglishProblem}
For several datasets, namely Math23k~\cite{Math23k}, Ape210k~\cite{Ape210k}, GaokaoBench-Math~\cite{GaokaoBench}, and AGIEval-Gaokao~\cite{AGIEval}, the problems are originally presented in Chinese. To ensure uniformity and mitigate the effects of multilingual representations, we translated these Chinese problems into English. The translation was facilitated by the GPT-3.5-Turbo API. Due to parsing errors encountered during the post-processing, a few examples were excluded. The prompt template employed for the translation request is provided below:

\texttt{\scriptsize
I want you to act as a Math Translator. Your task is to translate Chinese math questions into English math questions. \\
Make sure to keep the original question numbers. \\
Make sure to keep the math formula in Latex format. \\
The translations should be clear, accurate, and easily understandable for students who are native English speakers. \\
\\
\# Chinese Math Questions \#: \\
<insert chinese questions> \\
\\
\# English Math Questions \#:}

\subsection{CollegeMath: Extraction from textbooks}\label{appendix:Extract}

To construct the CollegeMath dataset, we made use of the GPT-3.5-Turbo API to parse and extract questions and answers from raw, segmented LaTeX exercises and their corresponding solutions.

\subsubsection{Extracting Questions from Exercises}

The primary goal was to convert raw, potentially unstructured questions from math textbooks into well-formulated LaTeX-formatted questions. Below is the prompt template we utilized for this extraction process:

\texttt{\scriptsize
I want you to act as a Math Parser. Your task is to convert raw messy questions from a math textbook into well-structured LaTeX-formatted questions. \\
\\
Please ensure to retain the original question numbers.\\
If needed, prepend the original instructions to the parsed questions to make them more comprehensible.\\
If needed, skip the broken questions.\\
\\
<insert demo>\\
\\
\#Raw Questions\#:\\
```\\
<insert a chapter of practice>\\
```\\
\\
\#Well-structured LaTeX-formatted Questions\#:}

\subsubsection{Extracting Answers from Solutions}

Similarly, for answers, our aim was to transform raw, messy answers from textbooks into clear, LaTeX-formatted answers. Here's the template for this task:

\texttt{\scriptsize
I want you to act as a Math Parser. Your task is to convert raw messy answers from a math textbook into well-structured LaTeX-formatted answers.\\
\\
Please ensure to retain the original answer numbers.\\
If needed, skip the broken answers.\\
\\
<insert demo>\\
\\
\#Raw Answers\#:\\
```\\
<insert a chapter of answer>\\
```\\
\\
\#Well-structured LaTeX-formatted Answers\#:
}

By employing the aforementioned prompt templates, we were able to extract a comprehensive set of questions and answers, thereby forming the foundation of the CollegeMath dataset.

\subsection{{\our{}}: Concrete Examples}
\label{appendix:concrete_examples}

\subsubsection{More Extracted Topics}
A set of 30 topics, randomly chosen, is listed below to illustrate the variety:

\texttt{\scriptsize
"Arithmetic operations"
"Word problem solving"
"Mathematics"
"Money and finance"
"Problem-solving strategies"
"Arithmetic"
"Multiplication"
"Proportions"
"Basic arithmetic operations"
"Conversion of units"
"Measurement and weight"
"Multiplication and addition"
"Budgeting"
"Basic arithmetic"
"Wages and overtime"
"Calculating earnings"
"Arithmetic Sequences"
"Exponential Growth"
"Financial calculations"
"Problem solving"
"Algebraic expressions"
"Economics"
"Time"
"Business and finance"
"Ratio and proportion"
"Problem-solving"
"Time calculations"
"Addition"
"Distance"
"Speed"
}

\subsubsection{More Extracted Knowledge Points}
Similarly, we provide a list of 30 knowledge points, chosen at random, to demonstrate the depth and breadth of content:

\texttt{\scriptsize
"Random selection of marbles"
"Definition and properties of dot product"
"Manipulation of complex numbers"
"Calculation of time required to complete a task"
"How to apply the concept of a seven-day cycle"
"Distinct numbers"
"Expectation of a function of a random variable"
"Ability to calculate total time"
"Combinations of numbers"
"Calculation of weekly income"
"Relative motion"
"Understanding the relationship between centimeters and kilometers"
"Diagonalizing a matrix"
"Proportional relationships between two quantities"
"Ergodic Markov chain"
"Addition of values"
"Counting the number of cars"
"Converting fractions to whole numbers"
"Identifying relationships between different variables"
"Ability to set up and solve a proportion equation"
"Addition and subtraction of matrices"
"Using logarithms to solve exponential equations"
"Probability of rolling a specific number on a six-sided die"
"Divisibility of polynomials"
"Application of multiplication to calculate total revenue"
"Identifying the highest and lowest scores"
"Ability to calculate percentages."
"Geometric interpretation of dot product"
"Dividing complex numbers"
"Understanding weight units"
}

\subsection{Evaluation  on Individual Topics}
\label{sec:topics}
We examine the subset performances on MATH, as shown in Table \ref{tab:MATH}. It is evident that \our{} consistently delivers exceptional results across diverse topics.

\begin{table*}[h]
\centering
\footnotesize
\setlength{\tabcolsep}{3pt}
\renewcommand{\arraystretch}{1.0}
\begin{tabular}{lccccccc}
\toprule
\multirow{2}{*}{\textbf{Model}} & \multicolumn{7}{c}{\textbf{MATH}}\\
\cmidrule{2-8}
& \makecell[l]{Prealgebra} & \makecell[l]{Algebra} & \makecell[l]{Intermediate\\Algebra} & \makecell[l]{Precalculus} & \makecell[l]{Probability} & \makecell[l]{Geometry} & \makecell[l]{Number\\Theory}\\
\midrule
\multicolumn{8}{c}{\textit{closed-source models}} \\
GPT-4 & \textbf{75.2} & \textbf{71.3} & \textbf{25.3} & \textbf{30.4} & \textbf{52.5} & \textbf{41.7} & \textbf{45.7} \\
GPT-3.5 & 59.3 & 55.5 & 17.3 & 20.1 & 30.1 & 29.8 & 30.3 \\
\midrule
\multicolumn{8}{c}{\textit{open-source models fine-tuned on LLaMA-2 13B}} \\
WizardMath & 23.6 & 21.4 & 7.5 & 7.1 & 10.9 & 12.3 & 6.8 \\
MAmmoTH & 21.4 & 17.2 & 6.9 & 7.8 & 11.8 & 8.7 & 6.2 \\
GAIR-Abel & 28.3 & 23.3 & 8.1 & 9.1 & 13.0 & 15.0 & 9.4 \\
MetaMath & 39.3 & 32.1 & 11.9 & 10.2 & 18.5 & 17.7 & 15.3 \\
\our{} & \textbf{52.9} & \textbf{53.4} & \textbf{13.6} & \textbf{17.3} & \textbf{24.6} & \textbf{25.6} & \textbf{25.7} \\
\midrule
\multicolumn{8}{c}{\textit{open-source models fine-tuned on LLaMA-2 7B}} \\
WizardMath & 16.5 & 15.2 & 6.3 & 5.8 & 6.7 & 8.5 & 5.9 \\
MAmmoTH & 15.1 & 12.5 & 6.5 & 4.3 & 9.9 & 7.3 & 6.1 \\
GAIR-Abel & 21.4 & 17.6 & 7.7 & 6.9 & 10.1 & 9.8 & 7.4 \\
MetaMath & 34.0 & 29.6 & 8.7 & 9.8 & 17.5 & 15.4 & 17.5 \\
\our{} & \textbf{48.9} & \textbf{49.3} & \textbf{12.4} & \textbf{15.2} & \textbf{23.2} & \textbf{23.3} & \textbf{23.8} \\
\midrule
\multicolumn{8}{c}{\textit{open-source models fine-tuned on Mistral 7B}} \\
\makecell[l]{WizardMath v1.1} & 51.4 & 50.7 & 13.9 & \textbf{19.9} & 25.5 & 24.4 & 22.4 \\
\makecell[l]{MetaMath Mistral} & 47.1 & 41.4 & 13.2 & 12.6 & 23.4 & 23.7 & 19.8 \\
\our{} & \textbf{55.9} & \textbf{52.8} & \textbf{14.6} & 18.6 & \textbf{28.9} & \textbf{26.5} & \textbf{27.5} \\
\bottomrule
\end{tabular}
\caption{Performance metrics across various topics on MATH. Within each section, the highest performing results are highlighted in bold font.}
\label{tab:MATH}
\end{table*}

We also detail the subset performances on CollegeMath at the College Level.  As shown in Table \ref{tab:CollegeMath}, despite the {\sc MwpBench} training set's seed questions only encompassing algebra, precalculus, and calculus, \our{} demonstrates robust performance in OOD's test sets including  vector calculus, probability, and linear algebra. However, an area of challenge is differential equations, where all models show limited success.

\begin{table*}[h]
\centering
\footnotesize
\setlength{\tabcolsep}{3pt}
\renewcommand{\arraystretch}{1.0}
\begin{tabular}{lccccccc}
\toprule
\multirow{2}{*}{\textbf{Model}} & \multicolumn{7}{c}{\textbf{CollegeMath}}\\
\cmidrule{2-8}
& \makecell[l]{Algebra} & \makecell[l]{Precalculus} & \makecell[l]{Calculus} & \makecell[l]{Vector\\Calculus} & \makecell[l]{Probability} & \makecell[l]{Linear\\Algebra} & \makecell[l]{Differential\\Equation} \\
\midrule
\multicolumn{8}{c}{\textit{closed-source models}} \\
GPT-4 & \textbf{41.1} & \textbf{21.2} & \textbf{20.6} & \textbf{29.0} & \textbf{11.5} & \textbf{6.5} & \textbf{1.2} \\
GPT-3.5 & 37.7 & 16.6 & 17.8 & 32.7 & 10.0 & 3.0 & 1.2 \\
\midrule
\multicolumn{8}{c}{\textit{open-source models fine-tuned on LLaMA-2 13B}} \\
WizardMath & 12.0 & 7.4 & 8.2 & 14.5 & 2.8 & 0.3 & 0.3 \\
MAmmoTH & 11.2 & 4.2 & 7.0 & 8.1 & 2.8 & 1.5 & 0.0 \\
GAIR-Abel & 15.3 & 6.0 & 5.0 & 3.6 & 2.1 & 1.9 & 1.6 \\
MetaMath & 19.4 & 9.8 & 5.6 & 8.1 & 1.4 & 1.1 & 0.3 \\
\our{} & \textbf{35.0} & \textbf{17.8} & \textbf{15.8} & \textbf{24.5} & \textbf{7.9} & \textbf{5.0} & \textbf{1.9} \\
\midrule
\multicolumn{8}{c}{\textit{open-source models fine-tuned on LLaMA-2 7B}} \\
WizardMath & 9.7 & 5.2 & 10.2 & 11.8 & 1.4 & 1.1 & 0.3 \\
MAmmoTH & 9.5 & 4.8 & 7.0 & 10.0 & 2.1 & 3.4 & 0.0 \\
GAIR-Abel & 12.0 & 4.2 & 5.2 & 6.3 & 3.5 & 1.5 & \textbf{1.6} \\
MetaMath & 19.1 & 6.8 & 4.4 & 5.4 & 2.8 & 2.6 & 0.3 \\
\our{} & \textbf{34.2} & \textbf{19.6} & \textbf{18.8} & \textbf{27.2} & \textbf{7.9} & \textbf{5.0} & 0.6 \\
\midrule
\multicolumn{8}{c}{\textit{open-source models fine-tuned on Mistral 7B}} \\
\makecell[l]{WizardMath v1.1} & 29.3 & 14.0 & 11.4 & 16.3 & 5.0 & 2.3 & 0.0 \\
\makecell[l]{MetaMath Mistral} & 28.1 & 12.2 & 11.2 & 21.8 & 7.1 & \textbf{3.8} & 0.6 \\
\our{} & \textbf{37.1} & \textbf{18.0} & \textbf{19.4} & \textbf{27.2} & \textbf{8.6} & \textbf{3.8} & \textbf{1.6} \\
\bottomrule
\end{tabular}
\caption{Performance metrics across various topics on CollegeMath. Within each section, the highest performing results are highlighted in bold font.}
\label{tab:CollegeMath}
\end{table*}

\end{document}